# MDReg-Net: Multi-resolution diffeomorphic image registration using fully convolutional networks with deep self-supervision

Hongming Li, Yong Fan, and for the Alzheimer's Disease Neuroimaging Initiative[#]

Center for Biomedical Image Computing and Analytics (CBICA), Department of Radiology, Perelman School of Medicine, University of Pennsylvania, Philadelphia, PA, 19104, USA

**Abstract**

We present a diffeomorphic image registration algorithm to learn spatial transformations between pairs of images to be registered using fully convolutional networks (FCNs) under a self-supervised learning setting. The network is trained to estimate diffeomorphic spatial transformations between pairs of images by maximizing an image-wise similarity metric between fixed and warped moving images, similar to conventional image registration algorithms. It is implemented in a multi-resolution image registration framework to optimize and learn spatial transformations at different image resolutions jointly and incrementally with deep self-supervision in order to better handle large deformation between images. A spatial Gaussian smoothing kernel is integrated with the FCNs to yield sufficiently smooth deformation fields to achieve diffeomorphic image registration. Particularly, spatial transformations learned at coarser resolutions are utilized to warp the moving image, which is subsequently used for learning incremental transformations at finer resolutions. This procedure proceeds recursively to the full image resolution and the accumulated transformations serve as the final transformation to warp the moving image at the finest resolution. Experimental results for registering high resolution 3D structural brain magnetic resonance (MR) images have demonstrated that image registration networks trained by our method obtain robust, diffeomorphic image registration results within seconds with improved accuracy compared with state-of-the-art image registration algorithms.

[#] Data used in preparation of this article were obtained from the Alzheimer's Disease Neuroimaging Initiative (ADNI) database (adni.loni.usc.edu). As such, the investigators within the ADNI contributed to the design and implementation of ADNI and/or provided data but did not participate in analysis or writing of this report. A complete listing of ADNI investigators can be found at: http://adni.loni.usc.edu/wp-content/uploads/how_to_apply/ADNI_Acknowledgement_List.pdf

**Introduction**

Medical image registration plays an important role in many medical image analysis tasks (1, 2). To solve the medical image registration problem, the most commonly used strategy is to seek a spatial transformation that establishes pixel/voxel correspondence between a pair of fixed and moving images in an optimization framework, by maximizing a surrogate measure of the spatial correspondence between images, such as image intensity correlation between the images to be registered (3-7). Conventional medical image registration algorithms typically solve the image registration optimization problem using iterative optimization algorithms, making the medical image registration computationally expensive and time-consuming.

Recent medical image registration studies have leveraged deep learning techniques to improve the computational efficiency of conventional medical image registration algorithms, in addition to learning image features for the image registration using stacked autoencoders (8). In particular, deep learning techniques have been used to build prediction models of spatial transformations for achieving image registration under a supervised learning framework (9-12). Different from the conventional image registration algorithms, the deep learning based image registration algorithms formulate the image registration as a multi-output regression problem (9-12). They are designed to predict spatial relationship between image pixel/voxels from a pair of images based on their image patches. The learned prediction model can then be applied to images pixel/voxel-wisely to achieve the image registration.

The prediction based image registration algorithms typically adopt convolutional neural networks (CNNs) to learn informative image features and a mapping between the learned image features and spatial transformations that register images in a training dataset, consisting of deformation fields and images that can be registered by the deformation fields (9-12). Similar to most deep learning tasks, the quality of training data plays an important role in the prediction based image registration, and a variety of strategies have been proposed to build training data, specifically the spatial transformations that register images in a training dataset (9-12). Particularly, synthetic deformation fields can be simulated and applied to a set of images to generate new images so that the synthetic deformation fields can be used as training data to build a prediction model (11). However, the synthetic deformation fields may not effectively capture spatial correspondences between real images. Spatial transformations that register pairs of images can also be estimated using conventional image registration algorithms (9, 12). However, a prediction based image registration model built upon such a training dataset is limited to estimating spatial transformations captured by the adopted conventional image registration algorithms. The estimation of spatial transformations that register pairs of images can also be guided by shape matching (10). However, a large dataset of medical images with manual segmentation labels is often not available.

The training data scarcity problem in deep learning based image registration could be overcome using unsupervised or self-supervised learning techniques. A variety of deep learning algorithms have adopted deep CNNs, in conjunction with spatial transformer network (STN) (13), to learn prediction models for image registration of pairs of fixed and moving images in an unsupervised learning fashion (14-30). Particularly, fully convolutional networks (FCNs) that facilitate voxel-to-voxel learning (31) are adopted to predict the deformation field (14-17, 21, 24, 26, 28) using moving and fixed images as the input to deep learning networks. The optimization of the networks is driven by image similarity measures between the fixed image and the warped moving image based on either image intensity (14, 15, 17, 21, 24, 26, 28) or contextual features (16). The deformation field can be modeled by sufficiently smooth velocity fields to facilitate diffeomorphic image registration (6, 7), and such a strategy has been adopted in deep learning based image registration methods to favor diffeomorphic properties of the transformation including preservation of topology and invertible mapping (17, 25, 27). Although physically plausible deformation and promising accuracy has been obtained, these registration methods are carried out at a single spatial scale and might be trapped by local optima, especially when registering images with large anatomical variability. Inspired by conventional image registration methods, multi-stage and multi-resolution registration techniques are incorporated into deep learning based registration methods using cascaded networks (18, 20, 21, 26) and deep supervision (19, 20, 22, 23, 29), yielding improved performance compared with one-stage or single-scale image registration. However, they are not equipped to achieve diffeomorphic registration.

In this study, we propose an end-to-end learning framework to optimize and learn diffeomorphic spatial transformations between pairs of images to be registered in a **m**ulti-resolution **d**iffeomorphic image **reg**istration framework, referred to as MDReg-Net. In particular, our method trains FCNs to estimate voxel-to-voxel velocity fields of spatial transformations for registering images by maximizing their image-wise similarity metric, similar to conventional image registration algorithms. To account for potential large deformations between images, a multi-resolution strategy is adopted to jointly optimize and learn velocity fields for spatial transformations at different spatial resolutions progressively in an end-to-end learning framework. The velocity fields estimated at lower-resolution are used to warp the moving image and the warped moving image is used as the input to the subsequent sub-network to estimate the residual velocity fields for spatial transformations at higher-resolution. The image similarity measures between the fixed and warped moving images are evaluated at different image resolutions to serve as deep self-supervision so that FCNs at different spatial resolutions are jointly learned. A spatial Gaussian smoothing kernel is integrated with the FCNs to yield sufficiently smooth deformation fields to achieve diffeomorphic image registration. Our method has been evaluated based on 3D structural brain magnetic resonance (MR) images and obtained diffeomorphic image registration with better performance than state-of-the-art image registration algorithms.

**Methods**

**Image registration by optimizing an image similarity metric.** Given a pair of fixed image $I_f$ and moving image $I_m$, the task of image registration is to seek a spatial transformation that establishes pixel/voxel-wise spatial correspondence between the two images. The spatial correspondence can be gauged with a surrogate measure, such as an image intensity similarity measure between the fixed and transformed moving images, and therefore the image registration problem can be solved in an optimization framework by optimizing a spatial transformation that maximizes the image similarity measure between the fixed image and transformed moving image. For non-rigid image registration, the spatial transformation is often characterized by a dense deformation field $D$ that encodes displacement vectors between spatial coordinates of $I_f$ and their counterparts in $I_m$. For mono-modality image registration, mean squared intensity difference and normalized correlation coefficient (NCC) are often adopted as the surrogate measures of image similarity.

As the image registration problem is an ill-posed problem, regularization techniques are usually adopted in image registration algorithms to obtain a spatially smooth and physically plausible deformation field (1, 2). In general, the optimization based image registration problem is formulated as

$$\min_D -S\left(I_f(x), I_m(D \circ x)\right) + \lambda R(D), \qquad (1)$$

where $D$ is the deformation field to be optimized, $x$ represents spatial coordinates of pixel/voxels in $I_f$, $D \circ x$ represents deformed spatial coordinates of pixel/voxels by $D$ in $I_m$, $S(I_1, I_2)$ is an image similarity measure, $R(D)$ is a regularizer on the deformation field, and $\lambda$ controls the trade-off between the image similarity measure and the regularization on the deformation field.

The regularization is typically adopted to encourage the deformation field to be spatially smooth by minimizing magnitude of derivatives of the spatial transformation, such as square $L_2$-norm, total variation, and learning based regularizer (32, 33). To facilitate diffeomorphic image registration, the deformation field can be represented by integration of velocity fields $v$, i.e., $D = \Phi(v)$ (6, 7), and the regularization is directly applied to the velocity fields to obtain spatially smooth velocity fields and diffeomorphic deformation fields accordingly.

The image registration optimization problem can be solved by gradient descent based methods (1, 2). However, such an optimization based image registration task is typically computational expensive and time consuming. Instead of optimizing $D$ directly, the deformation field can be predicted using FCNs under an unsupervised setting (14, 15). However, the estimated deformation field may not be fold-free or invertible even a large smooth regularization term is adopted (17, 25, 27).

**Multi-resolution diffeomorphic image registration with deep self-supervision**. We adopt a multi-resolution image registration procedure to estimate the velocity and deformation fields progressively from coarse to fine spatial resolutions for its effectiveness for handling large deformation between images, as demonstrated in conventional image registration algorithms (1, 2). The overall framework of our multi-resolution image registration method is illustrated in Fig. 1a, with three different resolutions involved. Particularly, the velocity fields are estimated incrementally from coarse to fine resolutions with $L$ levels ($l = 1$ and $l = L$ refer to coarsest and finest spatial resolutions respectively), which are optimized jointly and formulated as

$$\min_{v^l} \sum_{l=1}^{L} -S\left(I_f^l(x), I_m^l(\Phi(\tilde{v}^l) \circ x)\right) + \lambda R(v^l), \qquad (2)$$

where $I_f^l$ and $I_m^l$ denote fixed and moving images at resolution level $l$, $v^l$ is the incremental velocity fields at level $l$, and $\tilde{v}^l$ is the accumulated velocity fields of the deformation field at level $l$, computed as

$$\tilde{v}^l = \sum_{i=1}^{l} v^i \text{ if } l > 1 \text{ and } \tilde{v}^1 = v^1. \qquad (3)$$

For the deformation field $\Phi(v^1)$ at the coarsest resolution ($l = 1$), a sub-network $S_1$ with a U-Net (34) architecture is utilized to estimate velocity fields $v^1$ and a moving image to be registered to a fixed image are concatenated with the fixed image as a two-channel input to sub-network $S_1$. For the deformation field at a finer resolution ($l > 1$), a dedicated sub-network $S_l$ is adopted to estimate the velocity field increment $v^l$ using a concatenation of the warped moving image and the fixed image as an input to the sub-network. Particularly, the moving image is warped by the deformation field $\Phi(\tilde{v}^l)$ obtained at its coarser resolution. The sub-network $S_l$ is optimized to learn the deformation field that captures the residual variation between the warped moving image and the fixed image after deformation at all preceding coarser resolutions. Finally, the accumulated velocity fields $\tilde{v}^L$ over all the resolutions are utilized to obtain the deformation field at the finest resolution.

Similar to the conventional multi-resolution image registration algorithms, the similarity of registered images at different resolutions is maximized in our network to serve as deep supervision (35), but without relying on any supervised information of the deformation fields. Such a supervised learning with surrogate supervised information is referred to as self-supervision in this study. As it is capable of obtaining both deformation and inverse deformation fields for the moving and fixed images from the velocity fields under the diffeomorphic image registration setting, our multi-resolution image registration model is formulated to optimize both the deformation and inverse deformation fields jointly

$$\min_{v^l} \sum_{l=1}^{L} -S\left(I_f^l(x), I_m^l(\Phi(\tilde{v}^l) \circ x)\right) - S\left(I_m^l(x), I_f^l(\Phi^-(\tilde{v}^l) \circ x)\right) + \lambda R(v^l), \qquad (4)$$

where normalized cross-correlation (NCC) is adopted as the image similarity measure $S(I_f, I_m)$, $R(v) = \sum_{n=1}^{N} \|\nabla v(n)\|_1$, $N$ is the number of pixel/voxels in the velocity field, and $\lambda$ is the hyper-parameter to balance the image similarity and deformation regularization terms.

Different from conventional multi-resolution image registration algorithms in which deformation fields at coarse resolutions are typically used as initialization inputs to the image registration at a finer resolution, our deep learning based method jointly optimizes deformation fields at all spatial resolutions with a typical feedforward and backpropagation based deep learning setting. As the optimization of the loss function proceeds, the parameters within the network will be updated through the feedforward computation and backpropagation procedure, leading to improved prediction of deformation fields.

**Network architecture for estimating the velocity fields.** In our multi-resolution image registration network, one dedicated sub-network is designed to estimate the velocity fields or the velocity field increment at each spatial resolution. The sub-network at the coarsest spatial resolution is optimized to learn the velocity fields to capture large deformation, while the sub-network at finer resolutions are optimized to learn residual deformation to achieve an accurate image registration.

In this study, stationary velocity fields (SVFs) $v$ are adopted to represent the deformation field as

$$\frac{\partial D^{(t)}}{\partial t} = v(D^{(t)}), \tag{5}$$

where $D$ is the deformation field, $D^{(0)} = Id$ is the identity transformation, and $t = [0,1]$ is time. The integration of SVFs $\Phi(v)$ using scaling and squaring method (6, 17) is adopted to compute the deformation field $D$ numerically. Particularly, the sub-network used at each resolution in our study is specified as one U-Net with both encoder and decoder paths, as illustrated in Fig. 1b. The encoder path of all the sub-networks share the same structure, consisting of one convolutional layer with 16 filters, followed by three convolutional layers with 32 filters, and all have a stride of 2. The decoder path of the sub-networks from the coarsest to finest resolutions have one, two, and three deconvolutional layers with 32 filters and a stride of 2 respectively, followed by two convolution layers with 32 and 16 filters respectively and one output convolutional layer to predict the SVFs at three different spatial resolutions. LeakyReLu activation is used for all the convolutional and deconvolutional layers except the output layer. The number of output channels $d$ is 3, corresponding to the spatial dimensionality of the input images. The kernel size in all layers are set to $3 \times 3 \times 3$. The multi-resolution images used for computing the image similarity in the loss function at different resolutions are obtained using average pooling. Specifically, the original image serves as the image at the finest (full) resolution, and images at reduced resolutions are obtained by applying average pooling to the original image recursively with a kernel size of $3 \times 3 \times 3$ and a stride of 2.

In the present study, SVFs are learned at $\frac{1}{8}$, $\frac{1}{4}$, and $\frac{1}{2}$ resolutions to reduce the computational memory consumption, and the SVFs at the full resolution are obtained from the output of the $\frac{1}{2}$ resolution using linear interpolation. The deformation field is computed from the SVFs with the number of time steps set to 7. A spatial smoothing layer, implemented as Gaussian kernel smoothing, is adopted as part of our deep learning network to smooth the network output at the finest resolution in the end-to-end learning framework as illustrated in Fig. 1a. Particularly, the $\sigma$ of the Gaussian kernel is set to 1.732 voxels and the kernel size is set to $3 \times 3 \times 3$. The integration of the spatial smoothing in our deep learning model facilitates adaptive learning of sufficiently smooth deformation fields to achieve the diffeomorphic image registration.

Our image registration model is implemented using Tensorflow (36). Adam optimization technique (37) is adopted to train the networks. Once the training procedure is finished, the trained network can be directly used to register new images with feedforward computation.

**Evaluation and experimental settings**

**Image datasets.** We evaluated our method based on two public brain imaging datasets with manual segmentations of fine-grained brain structures, including: 1) MICCAI 2012 Multi-Atlas Labelling Challenge (MALC) dataset consisting of T1 brain MR images from 30 subjects with fine-grained whole-brain annotation for 134 structures (38), and 2) Mindboggle-101 dataset consisting of T1 brain MR images from 101 healthy subjects with 50 manual annotated cortical structures (39). These images were used for validation only.

T1 brain MR images of 901 young subjects from PING dataset (40) were adopted to train our image registration model. Particular, images of 801 subjects were used for training, and images of the remaining 100 subjects were used for tuning the hyper-parameter $\lambda$. In addition, T1 brain MR images of 809 old subjects from ADNI 1 cohort (http://adni.loni.usc.edu) were adopted for training a second image registration model to investigate the influences of different training data to the registration performance.

All the images for model training and validation were preprocessed using FreeSurfer (41), including skull-stripping, intensity normalization and spatial alignment using affine registration. All the images were resampled with a spatial resolution of $1 \times 1 \times 1\ mm^3$ and cropped with a size of $176 \times 192 \times 176$. Segmentation labels with 30 brain structures were also obtained using FreeSurfer for each subject from PING dataset, which were adopted to tune the hyper-parameter $\lambda$.

**Evaluation metrics.** As it is non-trivial to obtain the ground truth deformation between any pair of images, we adopted the similarity of the anatomical segmentations of the fixed image and warped moving image as a surrogate metric of registration accuracy (42). Particularly, the trained registration model was applied to register all the testing images to one random selected template image, and the generated deformation fields were used to warp their corresponding segmentation labels. Dice score between the warped segmentation and the template segmentation images was used to evaluate the registration performance. Though Dice score between anatomical structures is a reliable surrogate measure to quantify image registration accuracy, higher Dice score alone does not necessarily mean biologically plausible image registration as a deformation field with folding voxels could also lead to image registration with high regional Dice score. Therefore, we also evaluated the diffeomorphic property of the obtained deformation in addition to Dice score. Particularly, we calculated the Jacobian determinant $|J_\Phi|$ of the deformation field $\Phi$ obtained and counted all the voxels $v$ whose $|J_\Phi(v)|$ is non-positive.

**Network training.** We trained pairwise registration models by randomly selecting one pair of images as the input to the network. Given a set of $n$ images, we obtained $n^2$ pairs of fixed and moving images, including pairs of the same images, such that every image can serve as the fixed image.

The learning rate was set to 0.0001 and batch size was set to 1. The networks were trained on one NVIDIA TITAN Xp GPU, and 150000 iterations were adopted for the training. We have trained our registration model with different hyper-parameter $\lambda$ values ($\lambda \in [0.1, 0.2, 0.35, 0.5, 0.75, 1]$) using the PING training dataset and selected the $\lambda$ values that obtained the highest Dice score on the PING validation dataset using the FreeSurfer segmentation labels while no voxels with non-positive $|J_\Phi(v)|$ existed in the obtained deformation fields.

**Comparison with state-of-the-art image registration algorithms and ablation studies.** We compared our method with representative medical image registration algorithms, including NiftyReg (43), ANTs (7), and VoxelMorph (44), based on the two testing datasets. Particularly, the default setting of NiftyReg was adopted. For ANTs based image registration, two configurations with different spatial smoothing regularization parameters were adopted with following command: ANTS 3 -m CC[fixed,moving,1,2] -t SyN[0.25] -r Gauss[9,0.2] (or -r Gauss[3,1.0]) -o output -i 201x201x201 --number-of-affine-iterations 100x100x100 --use-Histogram-Matching 0. The configuration with the small smoothing size is referred to as ANTs-c1, and the one with the larger smoothing size is referred to as ANTs-c2. For the VoxelMorph model, bi-directional image similarity based loss was adopted, and the number of time steps was set to 7 for computing the deformation field from the velocity field. The VoxelMorph model shared the same training strategy and setting as the proposed method, and its hyper-parameters were also optimized to obtain the highest Dice scores based on the PING validation dataset.

The comparison with VoxelMorph serves as an ablation study to evaluate if the multi-resolution strategy could improve the image registration. As an additional ablation study, we also investigated the performance of our method without the spatial smoothing layer by optimizing $\lambda$ to obtain the diffeomorphic image registration on the PING validation dataset.

**Experimental results**

**Optimal parameter setting.** Fig.2 shows the average Dice score and number of voxels with non-positive $|J_\Phi(v)|$ in the obtained deformation fields for the PING validation dataset with different values of hyper-parameter $\lambda$. It can be observed that the Dice scores reached the maximum when $\lambda$ was around 0.35, while all the voxels had positive $|J_\Phi(v)|$ in the obtained deformation fields when $\lambda$ was equal to or larger than 0.35. We adopted the registration model with $\lambda = 0.35$ for all the following evaluation unless specified otherwise.

**Quantitative performance of image registration algorithms under comparison.** The average Dice scores calculated over all anatomical structures and subjects obtained by different registration methods for two testing datasets are summarized in Table 1. All the deformable registration methods obtained significantly higher Dice scores than the affine image registration ($p < 4 \times 10^{-7}$, Wilcoxon signed rank test), and our method obtained deformation fields with the minimal numbers voxels with non-positive Jacobian determinant among all the methods under comparison. Our method also obtained Dice scores close to those obtained by ANTs-c1 and both of them ranked top in the deformable registration methods under comparison. Figs. 3 and 4 show Dice scores of individual anatomical structures of MALC and Mindboggle-101 datasets respectively, where the structures are presented in ascending order by their volumetric sizes (from small to large regions), and the Dice scores of the same anatomical structure from left and right brain hemispheres are combined. Our method was comparable to ANTs-c1 in terms of Dice score for most structures and outperformed the VoxelMorph model for most structures across both data sets with either coarse-grained (Mindboggle-101 dataset) or fine-grained (MALC dataset) structures. Example images before and after the image registration by different methods and their corresponding anatomical segmentations on two testing datasets are demonstrated in Fig. 5(a-d).

Example deformation fields and their corresponding Jacobian determinant maps for each dataset obtained by ANTs-c1, VoxelMorph, and our method are shown in Fig. 5(e and f). While there were several localized clusters of voxels with non-positive Jacobian determinant in the deformation fields obtained by ANTs-c1, nearly all voxels in the deformation fields obtained by VoxelMorph and our method were with positive Jacobian determinant, preserving good diffeomorphic property. As shown in Table 1, the average numbers of voxels with non-positive Jacobian determinant in the deformation fields obtained by ANTs-c1 were ~10 thousand, and that obtained by VoxelMorph were less than 20, while that obtained by our method were less than 0.2. These results indicate that incorporating the spatial smoothing layer in our method largely eliminated folding voxels in the deformation fields without much sacrificing registration accuracy. Though the folding voxels in the deformation fields obtained by ANTs could be eliminated by increasing the spatial smoothing during the registration, over-smoothing inevitably leads to degraded registration accuracy. As illustrated in Table 1, ANTs-c2 obtained image registration with a much smaller number of folding voxels compared with that obtained by ANTs-c1, but its Dice score decreased dramatically.

The average time used to register one pair of images by different registration methods are presented in Table 2. Our method and VoxelMorph took about 4.67 and 3.82 seconds respectively when run on an NIVIDIA TITAN Xp GPU, much faster than NiftyReg and ANTs which took about 257 and 1071 seconds on average when run on an Intel Xeon E5-2660 CPU.

As a deep learning based image registration model, the performance of the proposed method might be affected by the datasets used for training the image registration model due to the anatomical variations in different datasets. Therefore, we further trained image registration models using the proposed method and VoxelMorph on an image dataset from ADNI 1 cohort with the same training procedure as described previously and evaluated their performance on the two testing datasets. As summarized in Table 1, the image registration models trained on different datasets by our method had more stable and better image registration performance than those trained by VoxelMorph, demonstrating that our method is robust and capable of learning anatomical variations from different images.

Without the spatial smoothing layer, larger regularization parameter $\lambda$ was required to achieve diffeomorphic image registration. We trained image registration models without the spatial smoothing layer with different $\lambda$ values on the PING training dataset to identify the $\lambda$ value capable of generating deformation fields free of voxels with non-positive Jacobian determinant on the PING validation dataset. As shown in Fig. 6(a), $\lambda = 1.0$ produced an image registration model that registered the images of the PING validation dataset without any folding voxels while $\lambda = 0.5$ produced an image registration model that registered the images of the PING validation dataset with the maximal Dice score that was estimated based on the brain structures labeled by FreeSurfer. Fig. 6(b) shows numbers of voxels with non-positive Jacobian determinant of the ADNI1 images that were registered by the image registration models trained on the PING dataset with and without the spatial smoothing layer, respectively. Specifically, the average number of voxels with non-positive Jacobian determinant in the deformation fields obtained by MDReg-Net with the spatial smoothing layer was significantly less than that obtained by MDReg-Net without the spatial smoothing layer though a larger regularization parameter was used ($p = 3.59 \times 10^{-7}$, Wilcoxon signed rank test). At the subject level, the deformation fields of 31 out of 809 images obtained by MDReg-Net with the spatial smoothing layer contained voxels with non-positive Jacobian determinant, while 88 had deformation fields containing voxels with non-positive Jacobian determinant out of 809 images registered by MDReg-Net without the spatial smoothing layer. In terms of image registration accuracy measured by Dice scores on brain structures labeled by FreeSufer, MDReg-Net with and without the spatial smoothing layer obtained Dice scores of $0.782 \pm 0.115$ (mean $\pm$ standard deviation) and $0.777 \pm 0.116$, respectively ($p < 1 \times 10^{-10}$, Wilcoxon signed rank test).

The image registration accuracy of MDReg-Net image registration models with and without the spatial smoothing layer on the two testing datasets is summarized in Table 3. Particularly, two MDReg-Net image registration models without the spatial smoothing layer were obtained with $\lambda$ set to 0.5 and 1.0, respectively. Not surprisingly, MDReg-Net without the spatial smoothing layer could obtain better image registration accuracy than MDReg-Net with the spatial smoothing layer when $\lambda = 0.5$ at the cost of sacrificing the diffeomorphism. In contrast, MDReg-Net with the spatial smoothing layer could achieve diffeomorphic, albeit not perfect, image registration without sacrificing the image registration accuracy too much, compared with MDReg-Net without the spatial smoothing layer but with a larger regularization (when $\lambda = 1.0$).

**Discussion and Conclusions**

We present an end-to-end deep learning framework for diffeomorphic image registration. Our method trains FCNs to estimate voxel-to-voxel velocity fields of diffeomorphic spatial transformations for registering images by maximizing their image-wise similarity metric, similar to conventional image registration algorithms. To facilitate learning of large diffeomorphic deformations between images, a multi-resolution strategy is adopted to jointly optimize and estimate velocity fields of spatial transformations at different spatial resolutions incrementally with an integrated spatial Gaussian smoothing kernel. The experimental results based on 3D structural brain MR images have demonstrated that our method could obtain diffeomorphic image registration with better performance than state-of-the-art image registration algorithms.

Multi-stage and multi-resolution registration techniques have been incorporated into deep learning based registration methods using cascaded networks to account for large anatomical variations in recently studies (18, 20, 21, 26), and achieved improved accuracy compared with one-stage or single-scale registration. However, they are not equipped to achieve diffeomorphic registration. For deep learning based diffeomorphic registration methods, deep supervision has also been used to optimize image similarity at different spatial scales (22, 23). However, the deformations at different scales are learned separately, instead of being learned progressively. In our method, the velocity fields estimated at coarse resolutions are used to warp the moving image and the warped moving image is used as the input to the subsequent sub-network to estimate the residual velocity fields for spatial transformations at finer resolutions, so that the spatial transformations are learned from coarse to fine resolution incrementally. The image similarity measures between the fixed and warped moving images are evaluated at different image resolutions to serve as deep self-supervision so that FCNs at different spatial resolutions are jointly optimized.

We have evaluated our method using different brain structural image datasets with manually labeled anatomical segmentations available. These segmentations contains fine-grained anatomical structures, which are favored over brain tissue segmentation or coarse-grained segmentation for the evaluation of registration accuracy as suggested in literature (42). Given that high region overlap based accuracy (such as Dice score) does not necessarily indicate biological plausible deformations as folding voxels within regions could also result in high overlap index, we have also investigated the diffeomorphic property of the deformations obtained by different methods. As summarized in Table 1, our method obtained registration accuracy comparable to that obtained by ANTs, which is one top ranked diffeomorphic registration method, while our method obtained deformation fields with a much smaller number of folding voxels than those obtained by ANTs and other methods under comparison.

Our method obtained improved accuracy compared with VoxelMorph, which is a state-of-the-art deep learning based diffeomorphic registration model with similar deformation regularity and computational efficiency. This indicates that our incremental learning strategy could facilitate a better characterization of deformation between images. Compared with VoxelMorph, our method obtained more stable and accurate image registration models based on different brain image datasets with substantially different age distributions (younger than 20 vs. older than 60 years), indicating that our method is not sensitive to the training data though the age distributions of the subjects from the PING cohort and the ADNI cohort are different. As summarized in Table 2, deep learning methods on GPUs were much faster than conventional image registration algorithms on CPUs to register brain images.

Due to anatomical differences between images to be registered, the diffeomorphic image registration is often achieved at the cost of sacrificing the image registration accuracy in the current image registration framework which relies on regularization to produce spatially smooth and plausible deformation fields (1, 2). Although larger regularization parameters produced image registration models that could register images with smoother deformation fields, those producing image registration models to achieve the diffeomorphic image registration for the training data did not necessarily yield diffeomorphic image registration for the testing data and the discrepancy was prominent for the models trained without the spatial smoothing layer, as indicated by the results shown in Fig. 2 and Fig. 6(a) as well as in Table 1 and Table 3. This is because the regularization parameter could adjust the network parameters during the network training to yield spatially smooth deformation fields but does not directly regularize the deformation fields for registering testing image pairs during inference. The regularization effect is likely to vanish when there exists large discrepancy in morphometry and appearance between the testing and training data. In contrast, the spatial smoothing layer always carries out the smoothing operation in the same way no matter when applied to training or testing images. As indicated by the results summarized in Table 1, the MDReg-Net model with the spatial smoothing layer trained on the PING dataset achieved perfect diffeomorphic image registration on the Mindboggle-101 dataset without sacrificing the image registration accuracy, compared with alternative state-of-the-art image registration algorithms, including ANTs and VoxelMorph. Compared with the MDReg-Net models without the spatial smoothing layer, the MDReg-Net models with the spatial smoothing layer achieved better image registration accuracy and close to perfect diffeomorphic image registration, as indicated by the results summarized in Table 3. All these results indicated that the spatial smoothing layer could enhance diffeomorphic image registration.

Diffeomorphic image registration is desired for accurately localizing cortical areas and it has been demonstrated that surface-based image registration methods achieved substantially better performance than conventional volume-based image registration methods (45). Our method provides an alternative means to achieve fast, accurate, and nearly perfect diffeomorphic brain image registration, facilitating computationally efficient brain image registration and brain mapping in large scale neuroimaging studies of brain development and neuropsychiatric disorders.

The present framework for diffeomorphic image registration could obtain image registration results within seconds with higher accuracy than state-of-the-art image registration algorithms without diffeomorphism violation, however, potential refinements in the following aspects may further improve the registration performance. First, the architecture and parameter setting of the networks used could be further optimized. Second, stationary velocity fields were adopted to model spatial transformations currently, which may have inferior performance for charactering large deformations that are needed in certain scenarios, such as modeling morphology of developing and aging brains. Using time-varying velocity fields (46) to model spatial transformations will be a good direction for future work. Finally, the regularization based image registration framework may be replaced with a constrained optimization framework to train a deep learning model with diffeomorphic image registration constraints to gain further improvement.

In summary, we have developed a deep learning method, referred to as MDReg-Net, for diffeomorphic image registration and experimental results have demonstrated MDReg-Net could obtain robust, diffeomorphic, albeit not perfect, brain image registration for different datasets.

**Acknowledgments**

This work was supported in part by National Institutes of Health grants [grant numbers EB022573, MH107703, DA039215, and DA039002]. We gratefully acknowledge the support of NVIDIA Corporation with the donation of the Titan GPUs used in this study.

Data collection and sharing for this project was funded by the Alzheimer's Disease Neuroimaging Initiative (ADNI) (National Institutes of Health Grant U01 AG024904) and DOD ADNI (Department of Defense award number W81XWH-12-2-0012). ADNI is funded by the National Institute on Aging, the National Institute of Biomedical Imaging and Bioengineering, and through generous contributions from the following: AbbVie, Alzheimer's Association; Alzheimer's Drug Discovery Foundation; Araclon Biotech; BioClinica, Inc.; Biogen; Bristol-Myers Squibb Company; CereSpir, Inc.; Cogstate; Eisai Inc.; Elan Pharmaceuticals, Inc.; Eli Lilly and Company; EuroImmun; F. Hoffmann-La Roche Ltd and its affiliated company Genentech, Inc.; Fujirebio; GE Healthcare; IXICO Ltd.; Janssen Alzheimer Immunotherapy Research & Development, LLC.; Johnson & Johnson Pharmaceutical Research & Development LLC.; Lumosity; Lundbeck; Merck & Co., Inc.; Meso Scale Diagnostics, LLC.; NeuroRx Research; Neurotrack Technologies; Novartis Pharmaceuticals Corporation; Pfizer Inc.; Piramal Imaging; Servier; Takeda Pharmaceutical Company; and Transition Therapeutics. The Canadian Institutes of Health Research is providing funds to support ADNI clinical sites in Canada. Private sector contributions are facilitated by the Foundation for the National Institutes of Health (www.fnih.org). The grantee organization is the Northern California Institute for Research and Education, and the study is coordinated by the Alzheimer's Therapeutic Research Institute at the University of Southern California. ADNI data are disseminated by the Laboratory for Neuro Imaging at the University of Southern California.

**Figures and Tables**

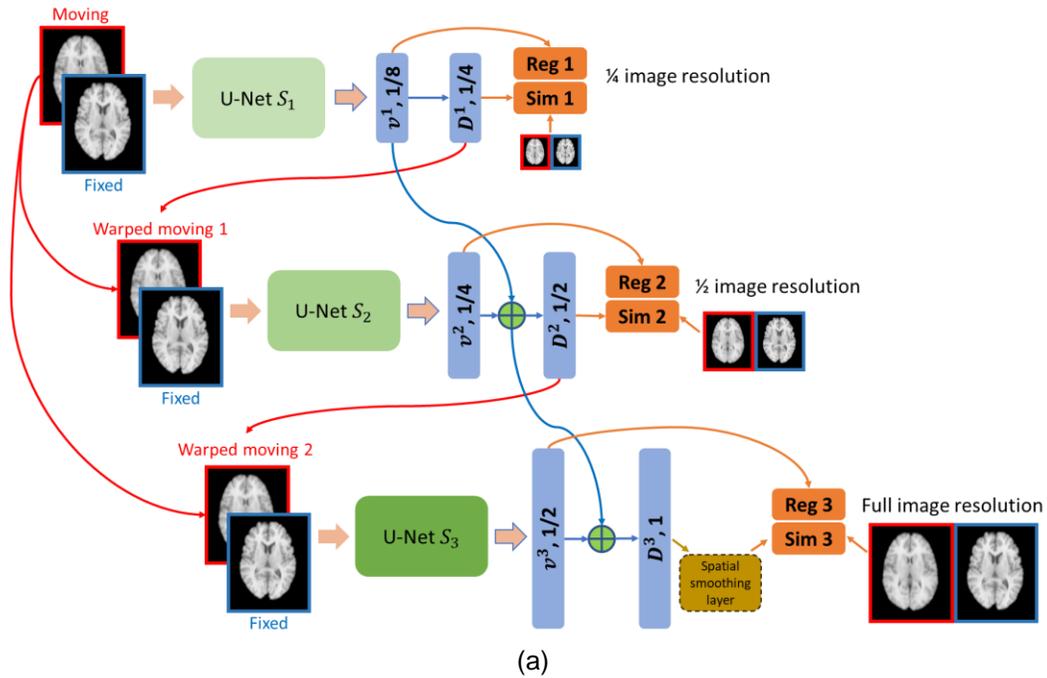

(a)

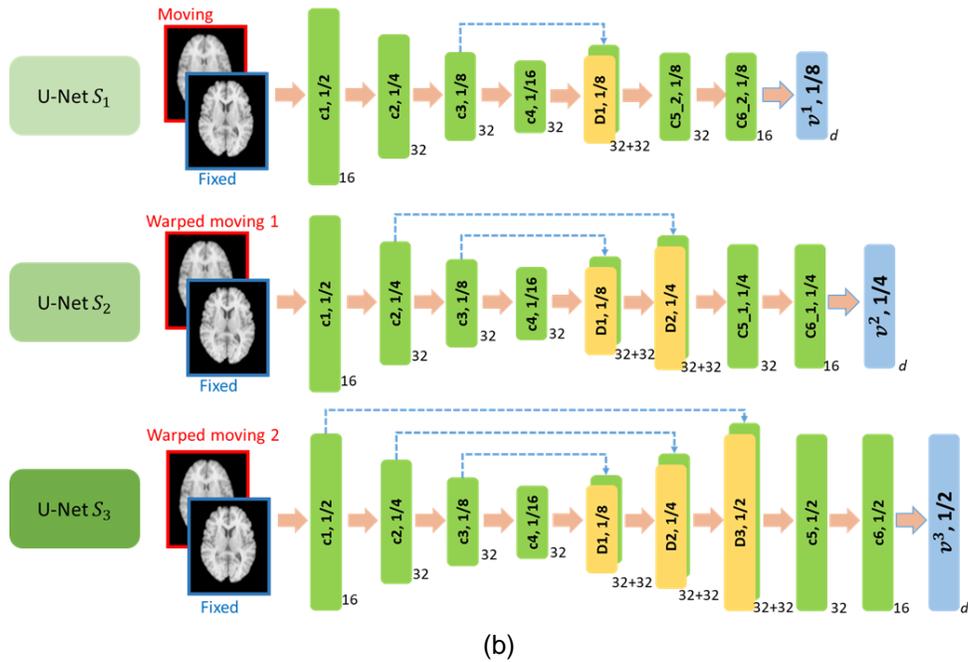

(b)

**Figure 1.** Schematic illustration of the multi-resolution diffeomorphic image registration based on FCNs. (a) Overall architecture of the multi-resolution image registration framework, (b) detailed network structure for voxel-to-voxel multi-output regression of velocity fields at different image resolutions. The number next to each network block denotes the number of its filters, and the number on each network block denotes the image resolution.

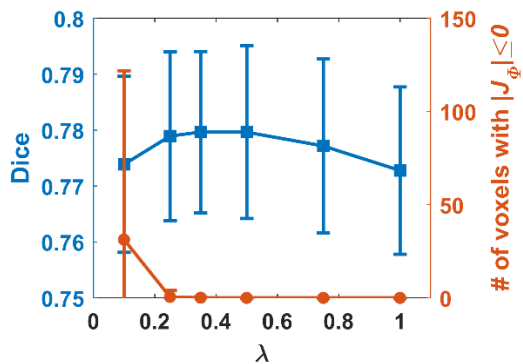

**Figure 2.** Dice score and number of voxels with non-positive $|J_\Phi(v)|$ in the obtained deformation fields of the PING validation dataset for the proposed model with different $\lambda$ values. The registration model with $\lambda = 0.35$ was adopted for all the evaluation.

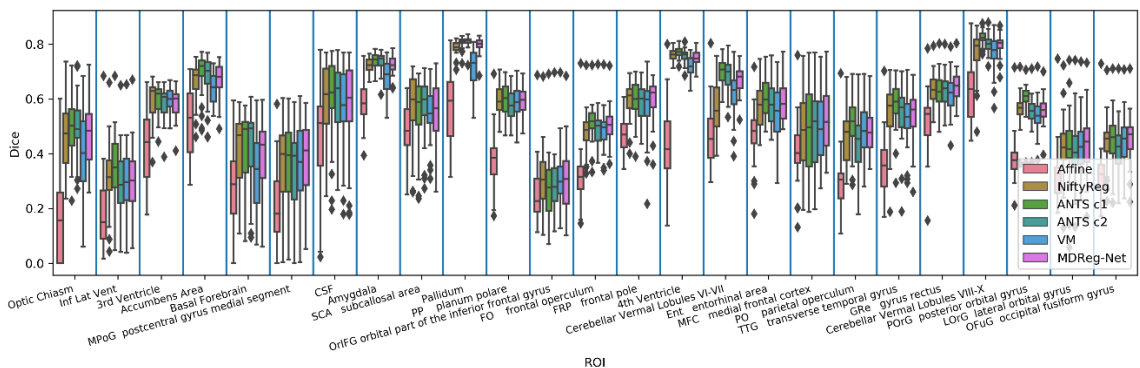

**Figure 3.** Boxplots of Dice score of anatomical structures for Affine, NiftyReg, ANTs (c1 and c2), VoxelMorph (VM), and our method (MDReg-Net) on the MALC dataset. Dice scores of the same structure from left and right brain hemispheres are combined. Brain structures are displayed in ascending order by their volumetric sizes from left to right.

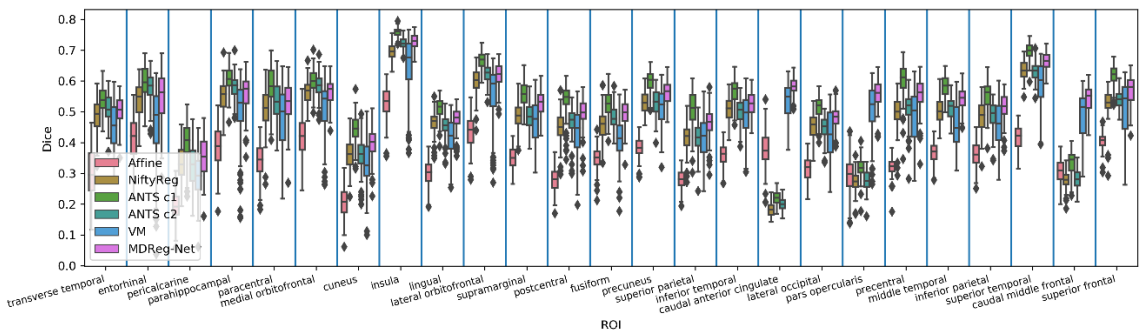

**Figure 4.** Boxplots of Dice score of 25 selected anatomical structures for Affine, NiftyReg, ANTs (c1 and c2), VoxelMorph (VM), and our method (MDReg-Net) on the Mindboggle-101 dataset. Brain structures are displayed in ascending order by their volumetric sizes from left to right.

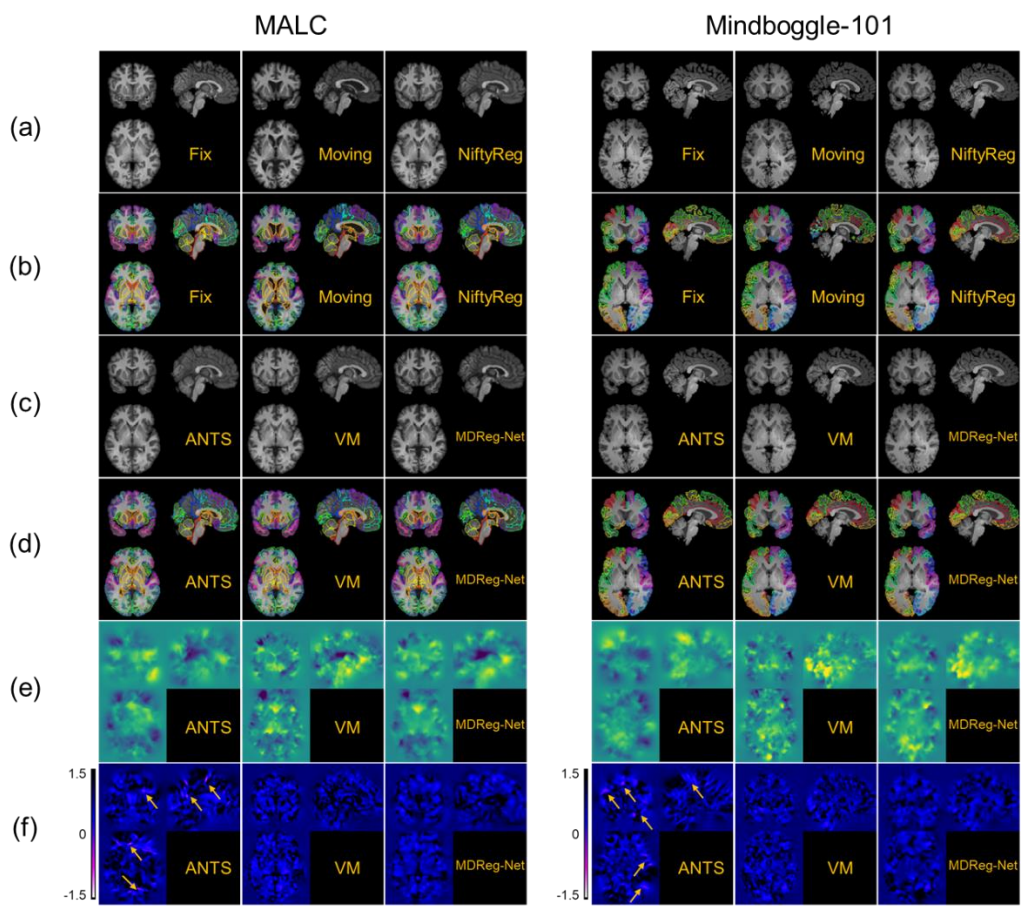

**Figure 5.** Example images before and after the image registration, obtained by the image registration algorithms under comparison on MALC and Mindboggle-101 datasets, respectively. (a and c) Fixed image, moving image, and warped moving images by NiftyReg, ANTs-c1, VoxelMorph (VM), and our method (MDReg-Net). (b and d) Segmentations of fixed and moving image, and warped segmentation of moving image by different registration methods. (e) Deformation fields obtained by ANTs-c1, VM and MDReg-Net to register the moving image to the fixed image. Deformation in each spatial dimension is mapped to one of the RGB color channels for the visualization. (f) Jacobian determinant maps of the deformation fields shown in (e). Localized clusters of voxels with non-positive Jacobian determinant are pointed out by the arrows. The average numbers of voxels with non-positive Jacobian determinant obtained by ANTs-c1, VM and MDReg-Net were 11945, 3.68, and 0.089 respectively on MALC dataset, while that were 11308, 12.39, and 0 respectively on Mindboggle-101 dataset.

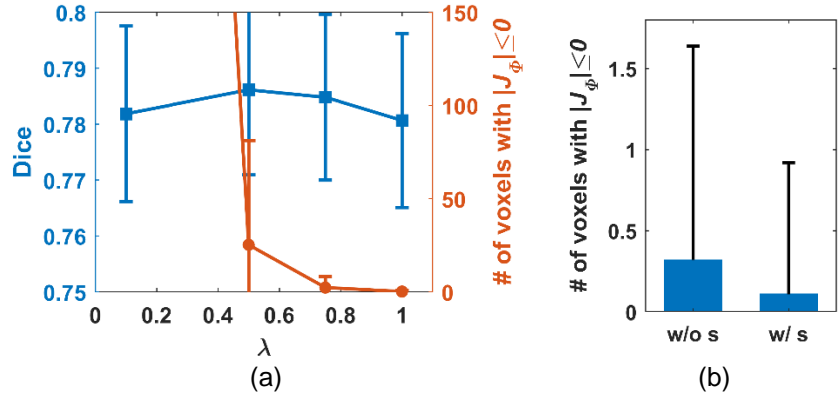

**Figure 6.** (a) Dice score and number of voxels with non-positive $|J_\Phi(v)|$ in the obtained deformation fields of the PING validation dataset by MDReg-Net without the spatial smoothing layer using different $\lambda$ values. The registration model with $\lambda = 1.0$ was adopted for the following ablation evaluation. (b) Number of voxels with non-positive $|J_\Phi(v)|$ in the obtained deformation fields of the ADNI 1 dataset using MDReg-Net without and with the spatial smoothing layer, respectively.

**Table 1.** Average Dice score and number of voxels with non-positive Jacobian determinant for affine alignment, NiftyReg, ANTs (SyN), VoxelMorph, and the proposed method (referred to as MDReg-Net) on different testing datasets. The performance of VoxelMorph and MDReg-Net trained using ADNI 1 dataset are also presented. The standard deviations are shown in parentheses.

|  | MALC | | Mindboggle-101 | |
|---|---|---|---|---|
| Methods | Avg. Dice | $|J_\Phi| \leq 0$ | Avg. Dice | $|J_\Phi| \leq 0$ |
| Affine | 0.429(0.182) | - | 0.347(0.093) | - |
| NiftyReg | 0.576(0.184) | - | 0.471(0.126) | - |
| ANTs-c1 | 0.597(0.187) | 11945(4448) | 0.538(0.130) | 11308(2788) |
| ANTs-c2 | 0.568(0.188) | 250.0(359.5) | 0.482(0.130) | 389.6(464.7) |
| VoxelMorph (PING) | 0.572(0.182) | 3.68(7.7) | 0.472(0.117) | 12.39(48.6) |
| VoxelMorph (ADNI1) | 0.568(0.182) | 4.53(9.8) | 0.476(0.113) | 8.72(38) |
| MDReg-Net (PING) | 0.588(0.180) | 0.089(0.515) | 0.534(0.094) | 0(0) |
| MDReg-Net (ADNI1) | 0.587(0.172) | 0.029(0.172) | 0.530(0.092) | 0.17(1.5) |

**Table 2.** Average runtime to register one pair of images by different registration methods. NiftyReg and ANTs run on one Intel Xeon E5-2660 CPU, while VoxelMorph and our method (MDReg-Net) run on one NVIDIA TITAN Xp GPU.

| Methods | NiftyReg | ANTs (SyN) | VoxelMorph | MDReg-Net |
|---|---|---|---|---|
| Avg. Time (sec) | 257 | 1071 | 3.82 | 4.67 |

**Table 3.** Average Dice score and number of voxels with non-positive Jacobian determinant for the proposed MDReg-Net without/with spatial smoothing layer on two testing datasets. The standard deviations are shown in parentheses.

|  | MALC | | Mindboggle-101 | |
|---|---|---|---|---|
| Methods | Avg. Dice | $|J_\Phi| \leq 0$ | Avg. Dice | $|J_\Phi| \leq 0$ |
| MDReg-Net w/o s ($\lambda = 0.5$, PING) | 0.592(0.181) | 19.85(21.9) | 0.539(0.099) | 6.45(9.3) |
| MDReg-Net w/o s ($\lambda = 0.5$, ADNI1) | 0.590(0.181) | 16.15(12.6) | 0.543(0.095) | 3.71(5.8) |
| MDReg-Net w/o s ($\lambda = 1.0$, PING) | 0.583(0.181) | 0.029(0.172) | 0.527(0.095) | 1.57(4.51) |
| MDReg-Net w/o s ($\lambda = 1.0$, ADNI1) | 0.580(0.180) | 0.088(0.515) | 0.523(0.094) | 0.01(0.10) |
| MDReg-Net (PING) | 0.588(0.180) | 0.089(0.515) | 0.534(0.094) | 0(0) |
| MDReg-Net (ADNI1) | 0.587(0.172) | 0.029(0.172) | 0.530(0.092) | 0.17(1.5) |